\documentclass{article}

\usepackage[numbers, compress]{natbib}

\usepackage[letterpaper]{geometry}
\usepackage{amsmath}        % aligned, align, \xrightarrow, etc.
\usepackage[utf8]{inputenc} % allow utf-8 input
\usepackage[T1]{fontenc}    % use 8-bit T1 fonts
\usepackage{hyperref}       % hyperlinks
\usepackage{url}            % simple URL typesetting
\usepackage{booktabs}       % professional-quality tables
\usepackage{amsfonts}       % blackboard math symbols
\usepackage{nicefrac}       % compact symbols for 1/2, etc.
\usepackage{microtype}      % microtypography
\usepackage{xcolor}         % colors
\usepackage{graphicx}       % include figures
\usepackage{placeins}       % keep floats within sections
\usepackage{multirow}       % multi-row cells in tables
\usepackage{tabularx}       % wrapping tables within the text block
\usepackage{CJKutf8}        % Chinese characters in case-study tables
\usepackage{authblk}         % multiple affiliations per author

\title{Joint Text-Audio Alignment for EEG-to-Text Decoding in Chinese Speech Production and Perception}

\author[1,2]{Tian Zheng\textsuperscript{*}}
\author[2]{Xurong Xie}
\author[3]{Xinxin Zhu}
\author[2]{Xiaolan Peng}
\author[2]{Feng Tian}

\affil[1]{University of the Chinese Academy of Sciences}
\affil[2]{Institute of Software, Chinese Academy of Sciences}
\affil[3]{Beijing Language and Culture University}
\affil[*]{E-mail: \texttt{zhengtian24@mails.ucas.ac.cn}}

\begin{document}

\maketitle

\begin{abstract}
Decoding speech information directly from scalp electroencephalography (EEG) into text provides a potential non-invasive neural communication pathway for individuals with severe speech and motor impairments. Compared with invasive approaches such as electrocorticography, EEG is safer and more widely deployable, but it remains substantially more challenging to decode. This challenge is exacerbated in Chinese sentence decoding, which must handle a high-dimensional output space with thousands of characters, severe inter-subject variability, and low signal-to-noise ratios for text alignment. Existing methods commit to a single supervisory axis---either text semantics or audio acoustic features---yet neither can simultaneously satisfy the demands of sentence-level discriminability and fine-grained temporal resolution required for large-vocabulary Chinese decoding. We introduce \textbf{EEGAlign}, a novel parameter-efficient framework that jointly aligns EEG along two axes: text alignment with BGE-M3 text embeddings and audio alignment with wav2vec~2.0 speech features via contrastive learning, followed by CTC character-sequence decoding. On ChineseEEG-2 data, EEGAlign yields state-of-the-art closed-set sentence classification performance, reaching up to $82.37\%$ Top-1 accuracy on Reading Aloud EEG and $41.43\%$ on Passive Listening EEG out of 101 candidates.Ablation studies show that the two alignment axes are highly complementary: combining them yields consistently better performance than either alone. To the best of our knowledge, this is the first study to decode large-vocabulary Chinese sentences from non-invasive EEG during overt speech production while achieving strong classification performance in a relatively large closed-set candidate-sentence setting.
\end{abstract}

\section{Introduction}

For individuals whose speech and motor functions are severely impaired by conditions such as locked-in syndrome, ALS, or stroke, brain--computer interfaces (BCIs) that decode speech intentions from neural activity offer a vital route toward restoring communication. While invasive recordings have demonstrated high-performance decoding~\cite{willett2023high}, the necessity of surgical implantation limits their broad clinical deployment. Scalp electroencephalography (EEG) presents a non-invasive, accessible, and safe alternative, yet it must overcome a significant ``quality gap'' characterized by low spatial resolution and three primary technical hurdles. First, \emph{signal weakness} requires extracting speech and language activity from noisy recordings affected by artifacts and background neural noise. Second, \emph{subject variability} in cortical anatomy and neurophysiology creates distribution shifts that hinder model generalization~\cite{jayaram2016transfer,ma2019reducing}. Finally, \emph{output-space complexity} makes mapping neural representations to discrete tokens particularly difficult. These issues are further exacerbated in Chinese sentence-level decoding, making it a major challenge for translational neural decoding.

Recent progress in EEG-to-text has addressed these challenges along three empirical lines, though most advances target passive listening or silent reading and evaluate open-ended generation metrics (BLEU/ROUGE) rather than exact-match accuracy. The first line integrates pretrained language models~\cite{duan2023dewave,wang2024enhancing,liu2024eeg2text} or scales to LLM backbones with semantic reranking~\cite{jiang2024neurolm,zheng2025guiding,hmamouche2026braindec}. While these methods gain sentence-level discriminability from powerful text priors, they share a critical limitation: alignment is purely semantic, providing no fine-grained temporal supervision---and their autoregressive decoders are prone to hallucination when EEG input is noisy. The second line reconstructs acoustic intermediates such as mel-spectrograms~\cite{fan2025ssm2mel,fan2025dmf2mel,xu2024convconcatnet}, motivated by the observation that direct text discretization discards prosodic structure still reflected in neural activity. These systems capture temporal dynamics but treat acoustic reconstruction as the main goal and rarely consider sentence-level discriminability. The third line builds foundation-scale EEG encoders via self-supervised pretraining~\cite{yang2023biot,jiang2024large}, offering generic representations that still require task-specific alignment for downstream decoding. Crucially, all three paradigms have been driven primarily by non-Chinese datasets (e.g., ZuCo~\cite{hollenstein2018zuco}), and none jointly exploits text and audio supervision.

The few studies investigating overt speech production typically restrict decoding to isolated syllables or words~\cite{li2023effects,ma20253m}. EEG-to-text decoding remains largely unexplored for continuous active speech; we found only one preprint~\cite{sato2024scalinglawneuraldata} using 175 hours of data and reporting relatively weak classification. In this setting, both articulation-induced motor signals and linguistic representations are crucial for decoding. ChineseEEG-2~\cite{chen2025eeg} provides the first Chinese continuous speech EEG dataset with matched Reading Aloud (RA) and Passive Listening (PL) paradigms. Yet no existing method jointly exploits text and audio supervision on this data: text semantic alignment alone lacks fine-grained temporal resolution, while audio acoustic alignment alone offers weak inter-sentence separability. Combining them addresses complementary granularities of speech and language representation.

Building on this observation, we propose \textbf{EEGAlign}, a parameter-efficient framework that jointly optimizes an EEG encoder with text alignment (to frozen BGE-M3 text embeddings) and audio alignment (to wav2vec~2.0 speech features) via contrastive learning alongside CTC character-sequence decoding. Lightweight bottleneck adapters ($\sim$16K params per subject) handle inter-subject variability, and a three-stage curriculum stabilizes multi-objective optimization. Experiments conducted on ChineseEEG-2 demonstrate the effectiveness of EEGAlign. The framework is backbone-agnostic: both a 9.91M-parameter Conformer and a pretrained LaBraM encoder~\cite{jiang2024large} achieve strong performance, confirming that gains stem from the alignment formulation. Our contributions are: (i) we propose EEGAlign as a principled combination of complementary text-semantic and speech-acoustic supervision for EEG-to-text decoding in speech production and perception; (ii) EEGAlign achieves state-of-the-art closed-set sentence classification performance on ChineseEEG-2, reaching up to $82.37\%$ on RA and $41.43\%$ on PL (101 candidates); and (iii) we systematically analyze the alignment mechanisms through ablations, representational geometry, and candidate-set sensitivity. To the best of our knowledge, this is the first study to decode large-vocabulary Chinese sentences from non-invasive EEG during overt speech production while achieving strong classification performance in a relatively large closed-set candidate-sentence setting.

\section{Related work}
Non-invasive EEG decoding has transitioned from rudimentary discrete unit classification toward continuous language reconstruction and retrieval. Current frontiers in the field focus on decoding continuous language directly from neural signals, with technical evolution primarily bifurcating into two trajectories: acoustic reconstruction and cross-modal alignment.

Early investigations established the feasibility of linking non-invasive brain recordings to continuous speech via contrastive learning between speech representations and neural activity~\cite{defossez2023decoding}. Subsequent research has focused on the fine-grained reconstruction of acoustic features, such as mel-spectrograms, by refining sequence modeling through State Space Models (SSMs) and dynamic multiscale fusion~\cite{fan2025ssm2mel, xu2024convconcatnet, fan2025dmf2mel}. However, these methods center on acoustic fidelity, which introduces an objective mismatch with the ultimate goal of text identification. Furthermore, they often lack explicit sentence-level semantic alignment.
Conversely, another research path aims to establish direct associations between EEG signals and text or semantic representations. This is achieved through multi-task contrastive learning~\cite{jiang2024neurolm}, CLIP-style architectures~\cite{cao2025eeg}, or quantization-based discretization for language model integration~\cite{duan2023dewave}. Notably, Wang et al.~\cite{wang2024self} demonstrated in a match-mismatch classification task that the simultaneous utilization of self-supervised speech representations and contextual text embeddings effectively enhances performance. This provides preliminary evidence for the complementary value of multimodal (acoustic and semantic) features. Nevertheless, a fundamental disparity exists between classification tasks and continuous sentence decoding: while the former merely identifies correspondences between signals, the latter requires the reconstruction of precise, coherent linguistic sequences from inherently noisy and ambiguous neural activity.

The aforementioned progress has been predominantly driven by English benchmarks. The tonal nature of Mandarin and the scarcity of continuous spoken-EEG corpora pose significantly greater challenges for sentence-level decoding. Current non-invasive Chinese spoken-EEG research remains confined to sub-word units, such as monosyllabic classification~\cite{li2023effects} or pinyin-level production~\cite{ma20253m}. Consequently, decoding large-vocabulary continuous Chinese sentences remains a research gap.

Recent advancements in representation learning and subject adaptation provide the technical bedrock for addressing these complexities. Large-scale EEG foundation models, such as BIOT~\cite{yang2023biot} and LaBraM~\cite{jiang2024large}, have demonstrated that self-supervised pretraining can yield robust, transferable neural representations. For cross-modal bridging, InfoNCE-style contrastive learning has emerged as the dominant approach~\cite{oord2018representation}. To mitigate severe inter-subject variability, Parameter-Efficient Fine-Tuning (PEFT) techniques, such as LoRA~\cite{hu2022lora}, have been widely adopted.

\section{Method}
\label{sec:method}

\subsection{Problem formulation and CTC-based decoder}

Given a multichannel EEG recording $X \in \mathbb{R}^{C \times T}$ acquired at 250 Hz from $C=128$ scalp electrodes over a temporal window of length $T \leq 750$, our goal is to decode the corresponding Chinese character sequence $y=(y_1,y_2,\ldots,y_L)$, where each $y_i$ is one Chinese character and $L$ is the number of characters in the target sentence. Here $T$ denotes the number of EEG time samples, whereas $L$ denotes the output length; the two are not assumed to match, and their alignment is handled by CTC. This setting is highly challenging due to the extremely large output space, severe inter-subject variability, and the intrinsically weak alignment between scalp EEG and discrete text tokens.

Current approaches to EEG-to-text decoding generally fall into two paradigms: autoregressive sequence-to-sequence generation (often utilizing pretrained language models) and non-autoregressive direct alignment. While autoregressive models possess strong language priors, they are prone to ``hallucinations'' when driven by noisy, low-signal-to-noise-ratio EEG data, often generating fluent but semantically disconnected text. 

Therefore, we establish a robust Connectionist Temporal Classification (CTC) architecture as our output decoder. Unlike autoregressive decoders, CTC strictly enforces monotonic temporal alignment between the brain signal and the character sequence. Given an encoder $f_\theta$, the EEG signal is mapped to latent states $H=f_\theta(X)$, followed by a linear prediction head $g_{\mathrm{ctc}}$ and optimized with the CTC negative log-likelihood:

\begin{equation}
  \mathcal{L}_{\mathrm{ctc}}
  =
  -\log P_{\mathrm{ctc}}\!\left(y \mid g_{\mathrm{ctc}}(H)\right).
  \label{eq:ctc_objective}
\end{equation}

Although this decoder avoids autoregressive hallucinations and directly optimizes character-level alignment, it relies almost entirely on the weak temporal correspondence between noisy EEG frames and sparse text labels. Consequently, a pure CTC model often struggles to form globally discriminative representations necessary for accurate sentence-level identification.

\subsection{The EEGAlign framework}

To overcome the limitations of the CTC-only decoder, we propose \textbf{EEGAlign}. A natural extension would be to introduce sentence-level semantic supervision by aligning EEG representations with pretrained text embeddings. However, EEGAlign is built on the core view that semantic and acoustic supervision are complementary rather than interchangeable.

Sentence-level text semantic alignment provides a stable global target for distinguishing among thousands of Chinese sentences, whereas audio acoustic alignment supplies temporally structured information related to pronunciation, rhythm, and sub-second speech dynamics. Because the two supervision signals target different granularities of language representation, they are complementary rather than interchangeable. Accordingly, EEGAlign augments the CTC decoding path with two auxiliary alignment axes: semantic alignment to frozen BGE-M3 text embeddings and acoustic alignment to wav2vec~2.0 speech features.

\begin{figure}[!t]
\centering
\includegraphics[width=\linewidth]{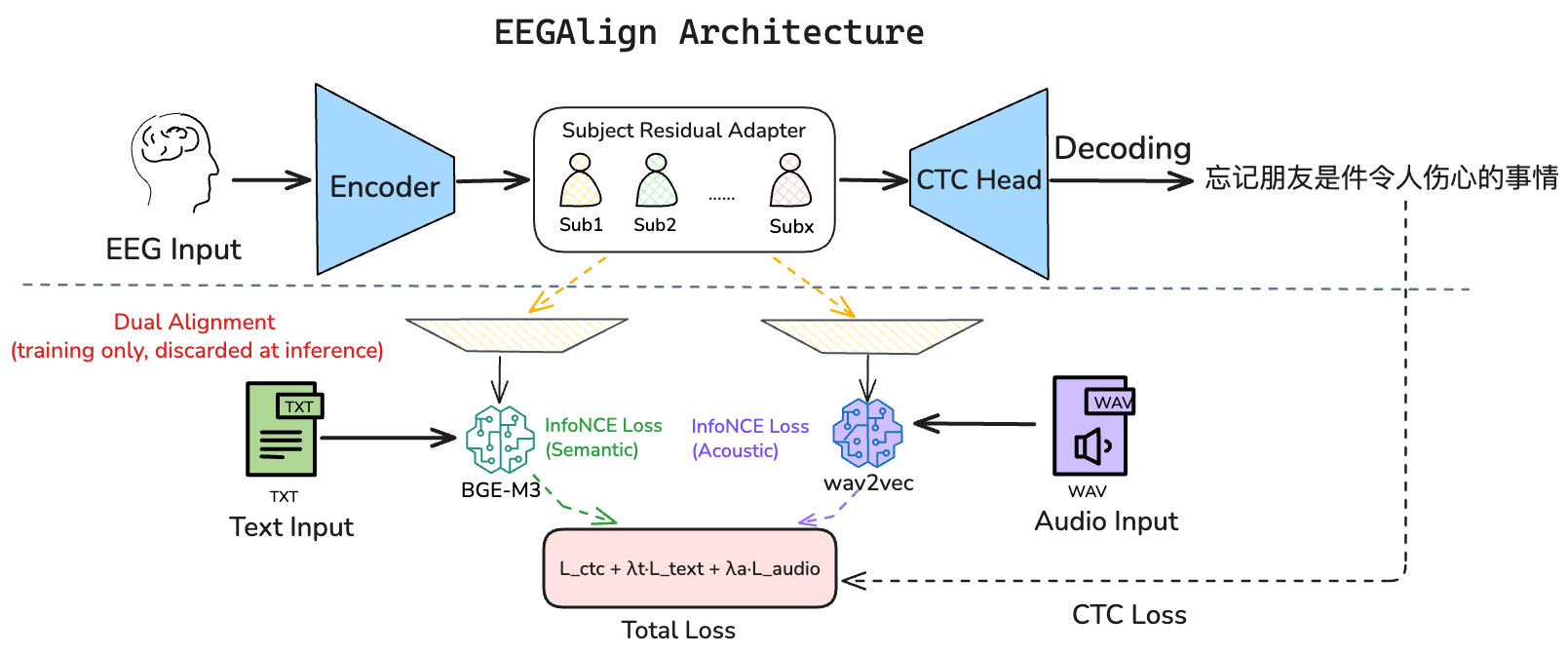}
\vspace{-3mm}
\caption{Overall architecture of \textbf{EEGAlign}. A shared sequence encoder with lightweight subject-specific adapters maps EEG to latent representations. A learned upsampler restores temporal density before applying three joint heads: CTC for decoding, text semantic alignment to BGE-M3 text embeddings, and audio acoustic alignment to wav2vec~2.0 features.}
\label{fig:main-architecture}
\vspace{-2mm}
\end{figure}

Figure~\ref{fig:main-architecture} provides an overview. Rather than treating the method as a rigid architecture, we organize it around one core requirement: the encoder must support \emph{three levels of supervision at once}---shared subject-agnostic structure, subject-specific adaptation, and cross-modal alignment at both sentence and chunk levels.

\subsection{Modeling with joint text-audio alignment}

We first describe EEGAlign at the level of its alignment and decoding objectives, treating the EEG encoder as an abstract module that produces a compact sequence representation $E \in \mathbb{R}^{B \times T_p \times d}$. This ordering emphasizes that the proposed text-audio alignment recipe is not tied to a particular backbone: any encoder that outputs a temporally ordered EEG representation can be coupled with the same CTC, text-alignment, and audio-alignment heads.

After downsampling in the shared encoder, the sequence has only about 30 time steps, which is sufficient for global discrimination but too coarse for frame-aligned character prediction and chunk-level acoustic matching. Rather than forcing CTC to operate directly on this compressed representation, EEGAlign upsamples the EEG encoder output to 50 Hz via a learned transposed convolution (kernel size and stride 5) followed by LayerNorm, matching the frame rate of wav2vec~2.0 audio features. This yields $\hat{E}\in\mathbb{R}^{B\times T_{\mathrm{up}}\times d}$ with $T_{\mathrm{up}}=T_p\times 5\approx 150$.

From this upsampled sequence, EEGAlign derives three complementary output paths: direct decoding, semantic alignment, and acoustic alignment. The direct decoding path linearly projects $\hat{E}$ to frame-level logits $P_{\mathrm{ctc}}\in\mathbb{R}^{B\times T_{\mathrm{up}}\times|\mathcal{V}|}$ and applies standard CTC loss~\cite{graves2006connectionist}. The semantic path mean-pools $\hat{E}$ into a sentence vector $\bar{e}_{\mathrm{eeg}}$, projects it into BGE-M3 space~\cite{chen2024bge}, and aligns it to the frozen text embedding of the ground-truth sentence via InfoNCE~\cite{oord2018representation} with temperature $\tau=0.05$:
\begin{equation}
  \mathcal{L}_{\mathrm{text}}=-\log\frac{\exp\bigl(\mathrm{sim}(\bar{e}_{\mathrm{eeg}},\bar{e}_{\mathrm{text}}^+)/\tau\bigr)}{\sum_j\exp\bigl(\mathrm{sim}(\bar{e}_{\mathrm{eeg}},\bar{e}_{\mathrm{text}}^j)/\tau\bigr)},
  \label{eq:text_align}
\end{equation}

The acoustic path uses frozen wav2vec~2.0 features~\cite{baevski2020wav2vec} as stable speech-side targets and projects the upsampled EEG sequence $\hat{E}$ into the same acoustic representation space. Both EEG and wav2vec~2.0 audio features are average-pooled into non-overlapping windows using \texttt{chunk\_size=20} frames (400~ms). We then perform chunk-level InfoNCE contrastive learning for time alignment, where the EEG chunk and audio chunk at the same temporal position form a positive pair, and all other chunks in the batch serve as negative samples.

Crucially, we employ a \textbf{symmetric contrastive formulation} to align noisy neural segments to stable speech features while keeping the speech-side representation fixed. We denote the directional chunk-level InfoNCE with query sequence $x$ and key sequence $y$ as
\begin{equation}
  \Phi(x,y)=-\frac{1}{M}\sum_m\log\frac{\exp\bigl(\mathrm{sim}(x_m,y_m^+)/\tau_a\bigr)}{\sum_j\exp\bigl(\mathrm{sim}(x_m,y_j)/\tau_a\bigr)}.
  \label{eq:phi}
\end{equation}
The bidirectional acoustic alignment loss is defined as
$\mathcal{L}_{\mathrm{audio}}^{\mathrm{e2a}}=\Phi(\hat{a},\mathrm{sg}(a))$ and
$\mathcal{L}_{\mathrm{audio}}^{\mathrm{a2e}}=\Phi(\mathrm{sg}(a),\hat{a})$,
where $\hat{a}_m$ and $a_m$ denote the EEG and wav2vec~2.0 audio chunk projections at position $m$, respectively, $\mathrm{sg}(\cdot)$ indicates stop-gradient, and $\tau_a=0.07$. Thus, gradients from the acoustic objective update the EEG-side representation, whereas the wav2vec~2.0 audio features remain fixed. The symmetric objective requires each EEG segment to identify its corresponding audio chunk and, conversely, to be uniquely discriminable when queried by the fixed audio anchor, thereby sharpening the representational structure against temporal smearing. The full objective is
\begin{equation}
  \mathcal{L}=\lambda_{\mathrm{ctc}}\mathcal{L}_{\mathrm{ctc}}+
  \lambda_{\mathrm{text}}\mathcal{L}_{\mathrm{text}}+
  \lambda_{\mathrm{audio}}\bigl(\mathcal{L}_{\mathrm{audio}}^{\mathrm{e2a}}+\mathcal{L}_{\mathrm{audio}}^{\mathrm{a2e}}\bigr),
  \label{eq:total_loss}
\end{equation}
where the contrastive formulation expresses supervision in terms of relative similarity. This is preferable to pointwise regression because EEG is noisy, temporally smeared, and amplitude-variable across subjects.

\paragraph{Inference-time candidate evaluation.}
Following related works~\cite{defossez2023decoding,wang2024self}, we evaluate a closed-set sentence classification task with a fixed set of text candidates. At inference time, each candidate $c$ is scored by fusing two complementary scores: (i) the \textbf{semantic score}, which computes the cosine similarity between the mean-pooled, $\ell_2$-normalized EEG projection $\mathbf{q}_{\mathrm{eeg}}$ and the pre-computed BGE-M3 text embedding $\mathbf{e}^{\mathrm{text}}_c$; and (ii) the \textbf{CTC score}, which computes the log-likelihood $\log P_{\mathrm{ctc}}(\mathbf{y}_c \mid X)$ of the candidate token sequence under the CTC head. Prior to fusion, both scores are calibrated by candidate-specific z-normalization using training-set statistics, denoted $z_c(\cdot)$. The final EEGAlign inference score is:
\begin{equation}
    s^{\mathrm{DA}}(X,c)
    =
    \alpha\, z_c\!\bigl(\cos(\mathbf{q}_{\mathrm{eeg}},\mathbf{e}^{\mathrm{text}}_c)\bigr)
    +
    (1-\alpha)\, z_c\!\bigl(\log P_{\mathrm{ctc}}(\mathbf{y}_c \mid X)\bigr),
    \label{eq:eval_fusion}
\end{equation}
with $\alpha=0.4$ by default. The candidate with the highest $s^{\mathrm{DA}}(X,c)$ is selected as the predicted sentence. 
\subsection{Encoder backbone choices and subject adaptation}

This section specifies the encoder module that supplies the compact representation $E$ consumed by the upsampler and the three supervision heads. EEGAlign is conceptualized as a flexible framework rather than being tied to a specific network architecture. The raw EEG signals are first mapped into temporal patches, and information is aggregated across channels and time to produce $E \in \mathbb{R}^{B \times T_p \times d}$. To demonstrate the generality of our text-audio alignment recipe, we instantiate the encoder using two distinct backbones:

\begin{itemize}
\vspace{-1mm}
    \item \textbf{Generic Encoder (Conformer):} A lightweight sequence encoder trained entirely from scratch. Each channel is processed independently by a strided 1D convolution (kernel and stride equal to 25), yielding $T_p=\lfloor T/25\rfloor$ patches of dimension $d=256$. Following learnable channel embeddings and Conditional Positional Encoding, the sequence is processed by 3 Conformer~\cite{gulati2020conformer} blocks. This 9.91M-parameter backbone establishes that dual alignment works effectively even without large-scale pretraining.
    \item \textbf{Foundation Encoder (LaBraM):} To test whether the framework scales with pretrained representations, we swap the generic encoder for LaBraM-base~\cite{jiang2024large}, a large-scale EEG foundation model. LaBraM retains explicit per-channel spatial information through its pretraining and maps the signals to a comparable latent space, requiring no changes to our alignment heads.
\vspace{-1mm}
\end{itemize}

A shared backbone may struggle to capture idiosyncratic subject patterns; however, per-subject encoder replication lacks the necessary parameter efficiency. We therefore inject subject adaptation only after the first two encoder layers, once a stable shared representation has begun to form. For subject $s$ at layer $l$, a bottleneck residual adapter~\cite{houlsby2019parameter} computes:
\begin{equation}
  \Delta E_l^{(s)}=\mathrm{SiLU}\bigl(E_l W_s^{\mathrm{down}}\bigr)W_s^{\mathrm{up}},
  \label{eq:adapter}
\end{equation}
where $W_s^{\mathrm{down}}\in\mathbb{R}^{d\times d_b}$, $W_s^{\mathrm{up}}\in\mathbb{R}^{d_b\times d}$, and $d_b=32$. We zero-initialize $W_s^{\mathrm{up}}$ so the adapter starts in the identity regime. The overhead is negligible (around 16K parameters per subject).

\subsection{Curriculum optimization}

Optimizing Equation~\ref{eq:total_loss} from scratch is challenging. 
Early in training, CTC optimization can be unstable because the model has not yet learned reliable token alignments over a large sequence space. Meanwhile, the contrastive objectives become informative only after the encoder has learned a minimally coherent representation space. We therefore train EEGAlign via a three-stage curriculum. We first perform a semantic 
warm-starting stage using only $\mathcal{L}_{\mathrm{text}}$, allowing the encoder to 
establish a basic latent structure before character-level supervision is introduced. 
We then enter a bootstrapping stage that activates CTC decoding as the primary objective 
while keeping the two alignment branches as auxiliary regularizers. Finally, we perform 
a joint refinement stage in which all three objectives remain active but the alignment 
terms are further down-weighted, so that CTC dominates the last phase of optimization 
without losing the representational structure established earlier. 

\section{Experiments}
\label{sec:experiments}

\subsection{Dataset}
\label{sec:dataset}

All experiments use \textbf{ChineseEEG-2}~\cite{chen2025eeg}, which extends ChineseEEG~\cite{mou2024chineseeeg} with active production and passive perception of speech over the same Chinese literary corpus. Materials consist of the Chinese translation of \emph{The Little Prince} (27~chapters, 1444~distinct Chinese characters) and excerpts from \emph{Garnett Dream} (8--9~recording sections, 2825~distinct Chinese characters). Twelve healthy right-handed participants (mean age 21.9; 4~male, 8~female) were recruited. In the \textbf{Reading Aloud} (RA) condition, four participants read the texts aloud while 128-channel EEG and speech were recorded, yielding about 10.8~hours of paired EEG--speech data. In the \textbf{Passive Listening} (PL) condition, the remaining eight participants listened to the RA recordings, yielding about 21.6~hours of EEG. The RA and PL groups are completely disjoint.

EEG was downsampled to 250~Hz, notch-filtered at 50~Hz, bandpass-filtered (1--40~Hz), cleaned with FastICA, and common-average re-referenced, following the default processing scripts. We re-segmented the raw streams to obtain sentence-level text targets and chunk-level wav2vec~2.0 acoustic targets~\cite{baevski2020wav2vec}, yielding examples with $30 \leq T \leq 750$. 

\textbf{Data Split Protocol.} For the main comparisons, both RA and PL use a within-subject stratified random split (90\% train, 10\% evaluation, seed = 42).  All main-text accuracies are computed on the held-out evaluation trials; Appendix~\ref{app:crossval} further reports a standard stratified five-fold cross-validation analysis on RA.
\subsection{Evaluation protocol and metrics}
\label{sec:evaluation}

We evaluate all methods using a sampled closed-set sentence identification protocol. For each paradigm, the candidate pool $\mathcal{C}$ contains all unique sentences in the dataset, tokenized into CTC targets $\mathbf{y}_i$. For each evaluation EEG trial (ground-truth $y^*$), we construct a candidate subset $\mathcal{C}_{\mathrm{sub}}$ containing one positive and $K$ (=100 by default) negative candidates sampled from $\mathcal{C} \setminus \{y^*\}$.

We report \textbf{Top-1}, \textbf{Top-5}, and \textbf{Top-10} accuracies, Character Error Rate (\textbf{CER}), and \textbf{CRA@1} (Chance-Ratio Accuracy at 1), defined as Top-1 accuracy divided by the random-chance baseline $1/(K+1)$; a CRA@1 of $n\times$ means the model is $n$ times better than chance.

\subsection{Experimental setup}

All models are trained on the same sentence-level splits and evaluated under the protocol described in Section~\ref{sec:evaluation}. This ensures that performance differences reflect model design rather than data splits or candidate construction.

Table~\ref{tab:impl} in Appendix~\ref{app:impl} summarises the implementation of EEGAlign. Unless otherwise stated, the encoder refers to the default lightweight instantiation described in Section~\ref{sec:method}; the alternative instantiation that replaces this encoder with a pretrained EEG foundation encoder shares all other components verbatim and is used as a second reference configuration.

\subsection{Baselines}

We compare EEGAlign against three external baselines chosen to reflect different strategies for strengthening EEG decoding. \textbf{LaBraM-MSE} uses the pretrained LaBraM EEG foundation model~\cite{jiang2024large} to encode EEG and trains a projection head with an MSE objective to match frozen BGE-M3 sentence embeddings. \textbf{DeWave}~\cite{duan2023dewave} is a representative EEG-to-text baseline based on discrete neural tokenization and a pretrained language model; in our implementation, its BART component is replaced with a Chinese-adapted BART checkpoint. \textbf{SMM-Challenge} adapts the EEG-Stimulus-Match-Mismatch model~\cite{wang2024self}, which combines self-supervised speech representations and contextual text embeddings for stimulus-response matching. All baselines are evaluated under our closed-set protocol using their own available ranking scores.\footnote{Resources for external baselines: \href{https://github.com/DeveloperSeJin/Dewave_Implementation}{DeWave implementation}, \href{https://huggingface.co/OpenMOSS-Team/bart-base-chinese}{Chinese BART checkpoint}, and \href{https://github.com/bobwangPKU/EEG-Stimulus-Match-Mismatch}{SMM-Challenge code}.}

\subsection{Main results}

We evaluate EEGAlign against three baselines using two distinct backbones: a lightweight Conformer and a pretrained LaBraM.
\begin{table}[htbp]
\caption{Closed-set sentence identification with one positive and 100 negatives.  }
\label{tab:closed_set}
\centering
\footnotesize
\setlength{\tabcolsep}{2.2pt}
\begin{tabular}{llrrrrrrrr}
\toprule
\multirow{2}{*}{Method} & \multirow{2}{*}{Encoder} & \multicolumn{4}{c}{RA (Reading Aloud)} & \multicolumn{4}{c}{PL (Passive Listening)} \\
\cmidrule(lr){3-6}\cmidrule(lr){7-10}
 & & Top-1 & Top-5 & Top-10 & CRA@1 & Top-1 & Top-5 & Top-10 & CRA@1 \\
\midrule
LaBraM-MSE~\cite{jiang2024large} & LaBraM-base & 23.33 & 45.58 & 56.92 & 23.57$\times$ & 19.44 & 43.84 & 57.02 & 19.64$\times$ \\
DeWave~\cite{duan2023dewave} & BART + VQ & 1.73 & 6.86 & 15.19 & 1.75$\times$ & 2.91 & 13.38 & 21.73 & 2.94$\times$\\
SMM-Challenge~\cite{wang2024self} & Brainnetwork & 26.22 & 70.90 & 89.55 & 26.48$\times$ & 3.11 & 9.57 & 17.00 & 3.14$\times$ \\
\midrule
\textbf{EEGAlign (ours)} & Conformer  & 69.62 & 91.15 & 94.04 & 70.31$\times$ & 27.26 & 63.72 & 78.13 & 27.54$\times$ \\
\textbf{EEGAlign (ours)} & LaBraM-base & \textbf{82.37} & \textbf{91.60} & \textbf{94.29} & \textbf{83.20$\times$} & \textbf{41.43} & \textbf{79.79} & \textbf{91.09} & \textbf{41.84$\times$} \\
\bottomrule
\end{tabular}

\vspace{-2mm}
\end{table}
\vspace{-1mm}

\noindent \textbf{Analysis of Results.} Table~\ref{tab:closed_set} presents the performance of \textit{EEGAlign}. We highlight two key insights:

\textit{1) Framework Effectiveness:} Both EEGAlign variants consistently outperform all baselines by a large margin. The foundation-encoder version achieves $82.37\%$ Top-1 on RA and $41.43\%$ on PL, demonstrating that dense auxiliary alignment to both text and audio is critical for robust decoding.

\textit{2) Encoder Robustness:} Substantial gains are observed across both the lightweight Conformer and the pretrained LaBraM backbone, confirming the improvements are framework-intrinsic. LaBraM's extra capacity to preserve channel-level spatial information further conditions the representation, yielding additional $+12.8$ and $+14.2$ point boosts on RA and PL.

Appendix~\ref{app:crossval} further reports a standard stratified five-fold RA cross-validation analysis, with a mean Top-1 accuracy of $82.20\%$ and a standard deviation of $1.68$ percentage points.

\subsection{Ablation studies}

Building on the main results, we conduct ablation studies to isolate the contributions of EEGAlign's individual components. All ablations use the 9.91M-parameter Conformer backbone and the same closed-set protocol with $K=100$ candidates as in Table~\ref{tab:closed_set}.  The variants in Table~\ref{tab:ablation} follow the table notation: \texttt{w/o audio} removes audio alignment, \texttt{w/o text} removes text alignment, \texttt{w/o both} removes both alignment objectives, \texttt{w/o adapter} removes subject-specific adapters, and \texttt{w/o curriculum} removes the staged training curriculum. 

\begin{table}[htbp]
\caption{Alignment-component ablation on the Conformer encoder.}
\label{tab:ablation}
\centering
\small
\setlength{\tabcolsep}{3.5pt}
\begin{tabular}{lrrrrrrrr}
\toprule
\multirow{2}{*}{Variant} & \multicolumn{4}{c}{RA (Reading Aloud)} & \multicolumn{4}{c}{PL (Passive Listening)} \\
\cmidrule(lr){2-5}\cmidrule(lr){6-9}
 & Top-1 & Top-5 & Top-10 & CER (\%)$\downarrow$ & Top-1 & Top-5 & Top-10 & CER (\%)$\downarrow$ \\
\midrule
\textbf{Full EEGAlign (text + audio)} & \textbf{69.62} & \textbf{91.15} & \textbf{94.04} & \textbf{28.90} & \textbf{27.26} & \textbf{63.72} & \textbf{78.13} & \textbf{68.30} \\
\quad w/o audio       & 56.92 & 79.55 & 82.05 & 59.80 & 21.12 & 58.62& \underline{74.71} & 71.00 \\
\quad w/o text       & 62.63 & 80.71 & 84.49 & 37.60 & \underline{26.92} & \underline{62.47} & 73.88 & \underline{64.70} \\
\quad w/o both         & 66.28 & 85.51 & \underline{88.59} & 34.30 & 23.55 & 59.92 & 70.60 & 67.10 \\
\quad w/o adapter    & 63.59 & 88.65 & 92.31 & 34.55 & 25.60 & 61.51 & 71.67 & 67.07 \\
\quad w/o curriculum  & \underline{67.05} & \underline{89.74} & 93.91 & \underline{31.30} & 25.54 & 60.47 & 72.08 & 67.67 \\
\bottomrule
\end{tabular}

\vspace{-2mm}
\end{table}
\vspace{-1mm}

Overall, Table~\ref{tab:ablation} shows that EEGAlign's gains come from three interacting factors: joint text--audio alignment, subject-specific adaptation, and progressive curriculum optimization. (i) The alignment objectives are complementary rather than independently additive. On RA, adding only one auxiliary axis is worse than using no auxiliary alignment: text-only alignment drops to $56.92$ and audio-only alignment reaches $62.63$, both below the CTC-only baseline ($66.28$). In contrast, combining the two axes improves Top-1 to $69.62$, the best result among all variants. This non-monotonic pattern suggests that either text or audio supervision alone may impose an incomplete constraint on the EEG encoder, while their joint use regularizes the representation from complementary perspectives. (ii) The subject-specific adapters are also important: removing them reduces Top-1 from $69.62$ to $63.59$ on RA and from $27.26$ to $25.60$ on PL. This indicates that the lightweight adapters help absorb inter-subject variability without replacing the shared encoder. (iii) Finally, removing the curriculum degrades both paradigms, indicating that progressive optimization is necessary to stabilize the heterogeneous CTC, text, and audio objectives.

\subsection{Analysis}
\label{sec:analysis}

This section provides a detailed analysis of the inference-time fusion gain, candidate-set sensitivity, and audio-alignment weight sensitivity. We further examine whether the semantic branch reshapes the encoder representation itself, with complementary paradigm-specific analyses deferred to Appendix~\ref{app:representation}.

\begin{figure}[htbp]
\centering
\vspace{-2mm}
\includegraphics[width=\linewidth]{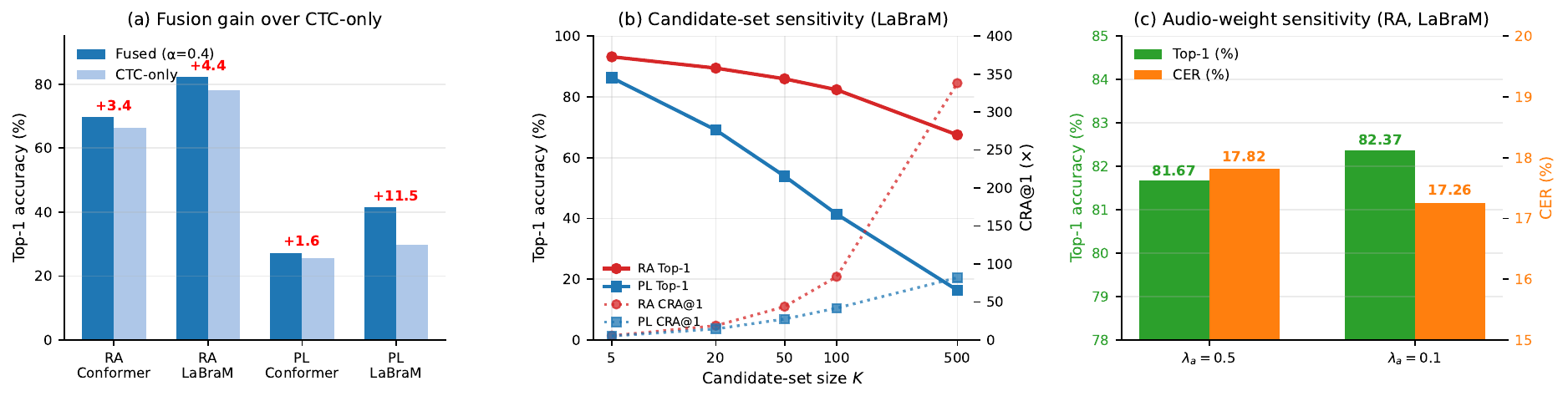}
\vspace{-3mm}
\caption{Analysis of fusion gain, hyperparameter sensitivity, and qualitative case studies. \textbf{(a)} Absolute Top-1 gain of fused score ($\alpha=0.4$) over CTC-only scoring; C and L denote Conformer and LaBraM. \textbf{(b)} Candidate-set sensitivity for EEGAlign-LaBraM with varying negative sizes $K$. \textbf{(c)} Sensitivity to the audio-alignment weight $\lambda_{\mathrm{audio}}$ on RA. \textbf{(d)} Qualitative UMAP visualisation: EEGAlign (red dots) moves representations of CTC-only baseline (blue dots) toward ground-truth BGE-M3 embeddings (stars).}
\label{fig:analysis_panel}
\end{figure}

\textbf{Fusion gain over CTC-only ranking.} We compare EEGAlign's default fused score $s^{\mathrm{DA}}$ (Eq.~\ref{eq:eval_fusion}, $\alpha=0.4$) to the degenerate case of ranking candidates by the CTC log-likelihood only, using the \emph{same} trained model. Figure~\ref{fig:analysis_panel}(a) shows the absolute Top-1 gain of fusion over CTC-only graphically (full numerical details are provided in Appendix~\ref{app:fusion_table}).

Two points are worth emphasising. (i) The fused score dominates the CTC-only score in every setting, confirming that the semantic embedding head and the CTC head provide complementary evidence at inference time. (ii) The magnitude of the gain depends strongly on paradigm and encoder: the largest gain ($+11.54$) is on PL with the LaBraM encoder, where CTC evidence is weak and the frozen BGE-M3 anchor is useful as a corrective signal (Table~\ref{tab:fusion}); the smallest gains ($+1.5$ to $+4.4$) are on RA and on the Conformer variants, where the CTC branch is already near-saturated. This is consistent with the view that text alignment functions mainly as a \emph{complementary inference-time score}, rather than as a replacement for the CTC score.

\textbf{Candidate-set sensitivity.} Because the absolute Top-1 accuracy of closed-set identification depends mechanically on the candidate-pool size $K+1$, we evaluate EEGAlign-LaBraM on both RA and PL under five candidate sizes ($K\in\{5,20,50,100,500\}$). As Figure~\ref{fig:analysis_panel}(b) summarises, RA Top-1 degrades gracefully from $93.21\%$ at $K{=}5$ to $67.50\%$ at $K{=}500$, while PL Top-1 drops from $86.32\%$ to $16.38\%$. Crucially, CRA@1 over chance \emph{grows} with $K$ for both paradigms. Although direct comparisons warrant caution due to differences in language and stimuli, our PL results generally align with—and, under our closed-set sentence-identification protocol, outperform—representative auditory speech-perception decoding and ICASSP challenge-style match-mismatch systems~\cite{defossez2023decoding,wang2024self}. Full per-$K$ results are reported in Appendix~\ref{app:ksens}.

\textbf{Audio-weight sensitivity.} We also examine the model's sensitivity to the audio-alignment weight $\lambda_{\mathrm{audio}}$ on RA. As shown in Figure~\ref{fig:analysis_panel}(c), the default gentler audio-alignment weight ($\lambda_{\mathrm{audio}}{=}0.1$) yields slightly better Top-1 accuracy ($82.37\%$ vs.~$81.67\%$) and lower Character Error Rate ($17.26\%$ vs.~$17.82\%$) compared to the stronger setting $\lambda_{\mathrm{audio}}{=}0.5$, indicating that while acoustic supervision is crucial, it should not overpower the primary CTC objective.

\textbf{Representation-level structural alignment.}
The fusion analysis above demonstrates that the semantic branch contributes useful inference-time evidence.  We compute Centered Kernel Alignment (CKA)~\cite{kornblith2019similarity} between the upsampled EEG encoder output ($d{=}256$, after the upsampler, before any projection head) and frozen BGE-M3 text embeddings ($d{=}1024$) on 200 held-out RA evaluation trials. The aggregate CKA results are reported in Appendix~\ref{app:cka}; in brief, EEGAlign achieves the highest structural alignment, increasing linear CKA from $0.294$ to $0.354$ and RBF CKA from $0.332$ to $0.402$ relative to the CTC-only baseline, while each single-axis variant improves CKA only mildly. This supports the same conclusion as Table~\ref{tab:ablation}.

Figure~\ref{fig:analysis_panel}(d) gives a compact qualitative view of this effect on representative rescued cases. The UMAP projection shows that EEGAlign moves the CTC-only baseline toward the correct BGE-M3 sentence anchor; the retained CKA summary and encoder-to-text representation measurements are provided in Appendices~\ref{app:cka} and~\ref{app:representation}.

\textbf{Case-level retrieval comparison.} Beyond the aggregate metrics, we inspect individual retrieval outcomes by examining the displayed case examples produced by EEGAlign, the CTC-only scoring baseline, and LaBraM-MSE (embedding-only retrieval without CTC). On 10 representative trials (5 RA, 5 PL) where both baselines fail at Top-1 but EEGAlign succeeds, the CTC-only baseline ranks the ground-truth sentence at positions 20--92 (median~$\sim$82) and LaBraM-MSE at positions 2--93, while EEGAlign places the correct sentence at rank~1 in every case. This indicates that text-audio alignment leverages complementary evidence: the alignment score and CTC log-likelihood provide strong sentence-level corrections to each other. Five cases for each paradigm and per-method rank comparisons are provided in Appendix~\ref{app:case_top5}.

% [Merged into Figure 2 as panel (d)]
\vspace{-3mm}

\section{Conclusion and limitations}
\vspace{-2mm}

EEGAlign investigates whether text and audio alignment provide complementary supervision for Chinese EEG-to-text decoding in speech production and perception. Our results suggest that simultaneous EEG--text and EEG--speech alignment yields strong closed-set classification performance under a 1-in-101 protocol and, crucially, does so across two distinct encoder instantiations. The generic Conformer and the pretrained LaBraM achieve up to $69.62\%$ and $82.37\%$ Top-1 on RA, respectively, outperforming all external baselines by large margins. This consistent gain confirms the robustness of the text-audio alignment recipe, independent of the specific backbone.

The two alignment supervision signals are not interchangeable: audio alignment acts as the dominant training-time auxiliary, while the text alignment head contributes primarily as a complementary inference-time score, with the largest fusion gain ($+11.5$ Top-1) on PL, where the sequence-level signal is weak. A gentle audio-alignment weight ($\lambda_{\mathrm{audio}}=0.1$) is sufficient and slightly preferable to a stronger one. Per-subject analysis (Appendix~\ref{app:persubject}) shows that all 12 subjects with reported values benefit substantially from full EEGAlign over CTC-only, with a tight spread across subjects (Top-1 std.~$\leq 2.3$ points on RA and $\leq 1.7$ points on PL for the best foundation-encoder variant). Together with the stratified five-fold RA cross-validation result (Top-1 $82.20\pm1.68\%$; Appendix~\ref{app:crossval}), these findings indicate that the proposed alignment recipe is stable under the present closed-set, subject-aware evaluation. This evaluation setting, however, is still a constrained step toward practical EEG-to-text communication: the current decoder scores a fixed candidate set rather than performing open-ended beam-search sentence decoding, and the subject-specific adapters require observed subject data rather than supporting zero-shot decoding for unseen users. More broadly, the modest number of subjects and limited amount of data also constrain the integration of EEG encoders with larger generative language decoders. Addressing these limitations is a key step toward deployable non-invasive Chinese neural communication systems.

\section*{References}

{\small
\renewcommand{\bibsection}{}
\bibliographystyle{unsrtnat}
\bibliography{reference}
}

\newpage
\appendix
\section{Comparison with representative neural decoding methods}
\label{app:related}

Table~\ref{tab:related} positions EEGAlign against representative methods across the design dimensions most relevant to our setting.

\begin{table}[htbp]
\caption{Representative neural decoding methods. CL: contrastive learning; PLM: pretrained language model; SR: Silent Reading; PL: Passive Listening; Overt: overt speech production.}
\label{tab:related}
\centering
\footnotesize
\setlength{\tabcolsep}{2pt}
\begin{tabularx}{\linewidth}{@{}>{\raggedright\arraybackslash}p{0.17\linewidth}ll>{\raggedright\arraybackslash}p{0.12\linewidth}>{\raggedright\arraybackslash}X>{\raggedright\arraybackslash}p{0.13\linewidth}>{\raggedright\arraybackslash}p{0.14\linewidth}@{}}
\toprule
Method & Venue & Lang. & Paradigm & Alignment strategy & Decoder & Task \\
\midrule
DeWave~\cite{duan2023dewave} & NeurIPS'23 & EN & SR & VQ $+$ PLM alignment & Pretrained LM & EEG $\rightarrow$ text \\
E2T-PTR~\cite{wang2024enhancing} & ACL'24 & EN & SR & CET-MAE transfer & BART decoder & EEG $\rightarrow$ text \\
NeuroLM~\cite{jiang2024neurolm} & ICLR'25 & mixed & multi-task & neural tokenizer $+$ instruction tuning & LLM backbone & EEG multi-task \\
LLM-Guided~\cite{zheng2025guiding} & ESWA & EN & SR & semantic EEG-language alignment & LLM-guided decoder & EEG $\rightarrow$ text \\
EEG2TEXT-CN~\cite{lu2025eeg2text} & arXiv'25 & CN & SR & text-EEG CL & MiniLM / dec. & EEG $\rightarrow$ text \\
D\'efossez et al.~\cite{defossez2023decoding} & NMI'23 & EN/Dutch & PL & wav2vec2 CL & Retrieval & MEG/EEG $\rightarrow$ speech segment \\
Wang et al.~\cite{wang2024self} & ICASSPW'24 & Dutch & PL & speech $+$ text CL & Match--mismatch & EEG--stimulus matching \\
ConvConcatNet~\cite{xu2024convconcatnet} & ICASSPW'24 & Dutch & PL-like & acoustic recon. & CNN & EEG $\rightarrow$ mel \\
SSM2Mel~\cite{fan2025ssm2mel} & ICASSP'25 & Dutch & PL-like & acoustic recon. & SSM/Mamba & EEG $\rightarrow$ mel \\
Li et al.~\cite{li2023effects} & NER'23 & CN & Overt & Riemannian feat. & LDA & EEG $\rightarrow$ tone/vowel \\
3M-CPSEED~\cite{ma20253m} & Sci Data'26 & CN & Overt, mouthed, imagined & Dataset & --- & Pinyin production \\

\midrule
\textbf{EEGAlign} & --- & \textbf{CN} & \textbf{Overt (RA) \& PL} & \textbf{Text $+$ audio} & \textbf{CTC head} & \textbf{EEG $\rightarrow$ text} \\
\bottomrule
\end{tabularx}
\end{table}

\section{Fused versus CTC-only scoring}
\label{app:fusion_table}
\begin{table}[htbp]
\caption{Scoring-protocol comparison on EEGAlign checkpoints ($K=100$). $\Delta$ is the absolute Top-1 gain of fused scoring over CTC-only.}
\label{tab:fusion}
\centering
\small
\begin{tabular}{llrrr}
\toprule
Setting & Encoder & Fused Top-1 (\%) & CTC-only Top-1 (\%) & $\Delta$ \\
\midrule
RA, Full EEGAlign                    & Conformer & 69.62 & 66.22 & $+3.40$ \\
RA, Full EEGAlign                    & LaBraM    & 82.37 & 78.01 & $+4.36$ \\
PL, Full EEGAlign                    & Conformer & 27.26 & 25.71 & $+1.55$ \\
\textbf{PL, Full EEGAlign}           & \textbf{LaBraM} & \textbf{41.43} & \textbf{29.89} & $\mathbf{+11.54}$ \\
\bottomrule
\end{tabular}
\end{table}

\section{Implementation and sensitivity details}
\label{app:details}

This appendix collects three items that are referenced from but not displayed in the main text: (i) the full implementation summary (Table~\ref{tab:impl}, Appendix~\ref{app:impl}), (ii) the full set of closed-set metrics across candidate-pool sizes $K\in\{5,20,50,100,500\}$ for EEGAlign-LaBraM on both RA and PL (Tables~\ref{tab:negative_sensitivity_ra}--\ref{tab:negative_sensitivity_pl}, Appendix~\ref{app:ksens}), and (iii) the complete results for the audio-alignment weight sweep (Table~\ref{tab:hyperparameter_sensitivity}, Appendix~\ref{app:lambda}). Each sub-appendix is a self-contained unit that reproduces the numbers that the main-text figures already show graphically.

\subsection{Implementation summary}
\label{app:impl}

Table~\ref{tab:impl} summarises the full implementation of EEGAlign used for every experiment in Section~\ref{sec:experiments}. Each run uses a single GPU and takes approximately 16 GPU-hours; peak memory is approximately 5GB for the Conformer instantiation and 10GB for the LaBraM-based instantiation. The reported experiments comprise roughly 20 independent runs, for an estimated total of about 320 GPU-hours; preliminary and debugging runs were modest relative to this total. The alignment heads, subject adapters, upsampler, CTC decoder, optimiser, and curriculum are held fixed across the two reference instantiations; only the encoder module differs.

\begin{table}[htbp]
\caption{Implementation summary of EEGAlign. Only the encoder module is swapped when comparing the two reference instantiations.}
\label{tab:impl}
\centering
\small
\begin{tabular}{@{}p{0.28\linewidth}p{0.66\linewidth}@{}}
\toprule
Component & Configuration \\
\midrule
Encoder (default, from scratch) & $L=3$ Conformer~\cite{gulati2020conformer}, $d=256$, $H=8$, $d_{\mathrm{ffn}}=512$, conv kernel $K=31$ \\
Encoder (alternative, pretrained) & LaBraM-base~\cite{jiang2024large}, frozen stem $+$ fine-tuned top blocks \\
Subject adapters & Bottleneck residual at layers 0--1, $d_b=32$, zero-init up-projection \\
Upsampler & ConvTranspose1D(kernel=stride=5) + LayerNorm \\
Text alignment & Linear$(d{\to}1024)$ $\to$ frozen BGE-M3, InfoNCE, $\tau=0.05$ \\
Audio alignment & Linear$(d{\to}768)$ $\to$ wav2vec~2.0, bidirectional InfoNCE, chunk $c=20$ \\
CTC decoder & Vocabulary $|\mathcal{V}|=21{,}128$, beam width 10 \\
Optimizer & AdamW, lr=$2\times10^{-4}$, weight decay $10^{-2}$, grad clip 5.0 \\
Batch size & 32 (PL) / 16 (RA); warmup = 4000; cosine annealing \\
Compute & Single GPU; ${\sim}$5GB peak memory for Conformer, ${\sim}$10GB for LaBraM, ${\sim}$16 GPU-hours per run \\
\textbf{Total parameters (default encoder)} & \textbf{${\sim}9.91$M} + 16.4K $\times$ subjects \\
\bottomrule
\end{tabular}
\end{table}

\subsection{Candidate-set sensitivity: full results}
\label{app:ksens}

Figure~\ref{fig:analysis_panel}(b) in the main text plots the Top-1 and CRA@1 columns of this experiment. Tables~\ref{tab:negative_sensitivity_ra}--\ref{tab:negative_sensitivity_pl} report the full metrics for the foundation-encoder (LaBraM) EEGAlign on RA and PL respectively.

\begin{table}[htbp]
\caption{Candidate-set sensitivity for closed-set retrieval (EEGAlign-LaBraM on RA, seed = 42). Random Top-1 is $1/(K{+}1)$.}
\label{tab:negative_sensitivity_ra}
\centering
\small
\begin{tabular}{rrrrrrrr}
\toprule
$K$ & Top-1 (\%) & Top-5 (\%) & Top-10 (\%) & CER (\%) & Random Top-1 (\%) & CRA@1 \\
\midrule
5   & 93.21 & 99.74 & 100.00 & 7.06  & 16.67 & 5.59$\times$ \\
20  & 89.49 & 96.67 & 98.97  & 10.87 & 4.76  & 18.79$\times$ \\
50  & 85.96 & 93.40 & 96.15  & 13.99 & 1.96  & 43.84$\times$ \\
100 & 82.37 & 91.60 & 94.29  & 17.26 & 0.99  & 83.20$\times$ \\
500 & 67.50 & 86.67 & 88.97  & 31.20 & 0.20  & 338.18$\times$ \\
\bottomrule
\end{tabular}
\end{table}

\begin{table}[htbp]
\caption{Candidate-set sensitivity for closed-set retrieval (EEGAlign-LaBraM on PL, seed = 42). Random Top-1 is $1/(K{+}1)$.}
\label{tab:negative_sensitivity_pl}
\centering
\small
\begin{tabular}{rrrrrrrr}
\toprule
$K$ & Top-1 (\%) & Top-5 (\%) & Top-10 (\%) & CER (\%) & Random Top-1 (\%) & CRA@1 \\
\midrule
5   & 86.32 & 100.00 & 100.00 & 15.32 & 16.67 & 5.18$\times$ \\
20  & 69.11 & 97.10  & 99.76  & 33.04 & 4.76  & 14.51$\times$ \\
50  & 53.90 & 89.50  & 96.30  & 48.65 & 1.96  & 27.49$\times$ \\
100 & 41.43 & 79.79  & 91.09  & 61.61 & 0.99  & 41.84$\times$ \\
500 & 16.38 & 49.03  & 65.10  & 85.71 & 0.20  & 82.06$\times$ \\
\bottomrule
\end{tabular}
\end{table}

\subsection{Audio-alignment weight sensitivity: full results}
\label{app:lambda}

Figure~\ref{fig:analysis_panel}(c) in the main text visualises the Top-1 and CER columns of this sweep. Table~\ref{tab:hyperparameter_sensitivity} reports the remaining metrics.

\begin{table}[htbp]
\caption{Audio-alignment weight sensitivity (foundation-encoder EEGAlign on RA).}
\label{tab:hyperparameter_sensitivity}
\centering
\small
\begin{tabular}{crrrrr}
\toprule
Audio weight $\lambda_{\mathrm{audio}}$ & Top-1 (\%) & Top-5 (\%) & Top-10 (\%) & CER (\%)$\downarrow$ & ctc\_top1 (\%) \\
\midrule
0.5 (stronger) & 81.67 & 92.12 & 94.74 & 17.82 & 78.97 \\
\textbf{0.1 (default)} & \textbf{82.37} & 91.60 & 94.29 & \textbf{17.26} & 78.01 \\
\bottomrule
\end{tabular}
\end{table}

\section{Per-subject results}
\label{app:persubject}

\begin{figure}[htbp]
\centering
\includegraphics[width=\linewidth]{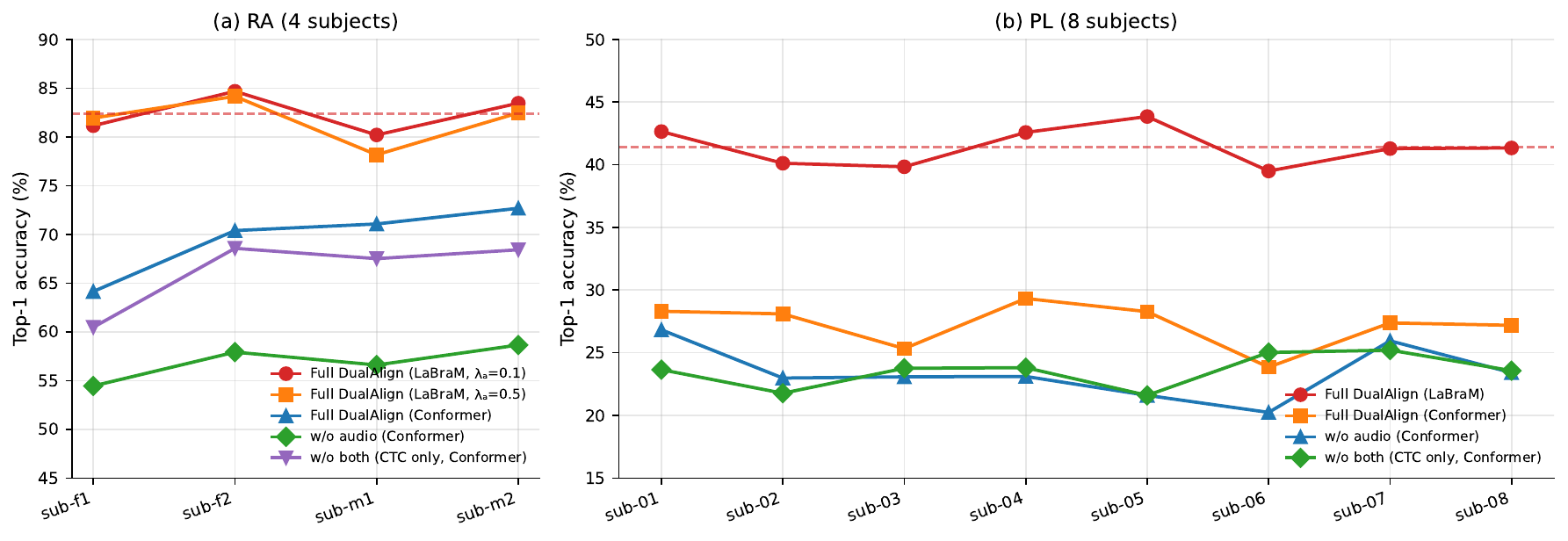}
\vspace{-3mm}
\caption{Per-subject Top-1 accuracy. \textbf{(a)} RA (4 subjects). \textbf{(b)} PL (8 subjects). Dashed lines mark the across-subject mean for EEGAlign-LaBraM.}
\label{fig:per_subject}
\vspace{-2mm}
\end{figure}

Tables~\ref{tab:persubject_ra}--\ref{tab:persubject_pl} and Figure~\ref{fig:per_subject} report the per-subject Top-1 accuracy (closed-set, $K=100$, seed = 42) at the best evaluation epoch for each training configuration, extracted directly from the training logs (the per-subject breakdown is emitted alongside every $\texttt{eval1/101}$ line as $\texttt{[eval][per\_subject] sub-X top1=\dots n=\dots}$). 

Three observations are worth highlighting. (i) The LaBraM + text-audio alignment model dominates all four RA subjects and all eight PL subjects uniformly: the subject-wise minimum on the best RA configuration ($\lambda_{\mathrm{audio}}{=}0.1$) is $80.20\%$ (sub-m1), still $+8$ points above the best Conformer subject, and the PL minimum is $39.50\%$ (sub-06), also above every Conformer subject. (ii) The per-subject spread is small (RA std.~$2.06$ points, PL std.~$1.67$ points for the best LaBraM variant), indicating that the reported aggregate accuracy is not driven by one or two outlier subjects. (iii) The Conformer ablations reveal a paradigm-dependent effect of the auxiliary objectives: on RA, the full model is best for every subject while CTC-only remains stronger than either single-axis variant; on PL, the audio-only variant is the second-best aggregate configuration and improves over CTC-only for every listed subject, consistent with the weaker sequence-level signal in passive listening.

\begin{table}[htbp]
\caption{Per-subject Top-1 (\%) on RA. $n$ is the number of evaluation trials per subject.}
\label{tab:persubject_ra}
\centering
\footnotesize
\setlength{\tabcolsep}{2.5pt}
\begin{tabularx}{\linewidth}{@{}>{\raggedright\arraybackslash}Xrrrrrr@{}}
\toprule
Configuration & sub-f1 & sub-f2 & sub-m1 & sub-m2 & Mean & Std \\
 & ($n{=}382$) & ($n{=}385$) & ($n{=}394$) & ($n{=}399$) & & \\
\midrule
\textbf{Full EEGAlign (LaBraM, $\lambda_{\mathrm{audio}}{=}0.1$)} & \textbf{81.15} & \textbf{84.68} & \textbf{80.20} & \textbf{83.46} & \textbf{82.37} & \textbf{2.06} \\
Full EEGAlign (LaBraM, $\lambda_{\mathrm{audio}}{=}0.5$) & 81.94 & 84.16 & 78.17 & 82.46 & 81.67 & 2.51 \\
Full EEGAlign (Conformer) & 64.14 & 70.39 & 71.07 & 72.68 & 69.62 & 3.75 \\
\quad w/o audio (Conformer)    & 54.45 & 57.92 & 56.60 & 58.65 & 56.92 & 1.86 \\
\quad w/o text (Conformer)    & 56.81 & 64.42 & 63.71 & 65.58 & 62.63 & 3.96 \\
\quad w/o both (Conformer)      & 60.47 & 68.57 & 67.51 & 68.42 & 66.28 & 3.87 \\
\bottomrule
\end{tabularx}
\end{table}

\begin{table}[htbp]
\caption{Per-subject Top-1 (\%) on PL. $n$ is the number of evaluation trials per subject.}
\label{tab:persubject_pl}
\centering
\footnotesize
\setlength{\tabcolsep}{1.5pt}
\begin{tabular}{lrrrrrrrrrr}
\toprule
Configuration & sub-01 & sub-02 & sub-03 & sub-04 & sub-05 & sub-06 & sub-07 & sub-08 & Mean & Std \\
\midrule
\textbf{Full EEGAlign (LaBraM)} & \textbf{42.65} & \textbf{40.12} & \textbf{39.84} & \textbf{42.58} & \textbf{43.85} & \textbf{39.50} & \textbf{41.29} & \textbf{41.34} & \textbf{41.43} & \textbf{1.57} \\
Full EEGAlign (Conformer)       & 28.31 & 28.10 & 25.33 & 29.33 & 28.28 & 24.16 & 27.38 & 27.19 & 27.26 & 1.79 \\
\quad w/o audio ( Conformer) &22.71 & 21.31 & 20.78 & 20.79 & 20.23 & 19.21& 22.66 & 21.27 & 21.12 & 1.16 \\
\quad w/o text ( Conformer) & 25.37 & 28.85 & 28.80 & 28.13 & 25.90 & 25.73 & 25.47 & 26.50 & 26.92 & 1.56 \\
\quad w/o both (Conformer) & 23.65 & 21.77 & 23.76 & 23.81 & 21.60 & 25.02 & 25.21 & 23.57 & 23.55 & 1.31 \\
\bottomrule
\end{tabular}

\vspace{3pt}

\end{table}

\FloatBarrier
\section{Five-fold cross-validation}
\label{app:crossval}

All main-text results (Tables~\ref{tab:closed_set}--\ref{tab:ablation}) use a single fixed 90/10 split (seed = 42). To assess split sensitivity, we additionally run standard stratified five-fold cross-validation for the LaBraM-based RA model. The five evaluation folds are mutually exclusive and together cover the RA trials; in each fold, 80\% of the data are used for training and the remaining 20\% for evaluation.

Table~\ref{tab:cv_perfold_ra} reports the per-fold closed-set metrics. The single fixed-split result from Table~\ref{tab:closed_set} is reproduced in the last row for reference, but should not be interpreted as one of the five cross-validation folds.

\begin{table}[htbp]
\caption{Five-fold cross-validation results of EEGAlign-LaBraM on RA.}
\label{tab:cv_perfold_ra}
\centering
\small
\begin{tabular}{lcccc}
\toprule
 & Top-1 (\%) & Top-5 (\%) & Top-10 (\%) & CER (\%)$\downarrow$ \\
\midrule
Fold 1 & 80.55 & 89.10 & 95.20 & 16.00 \\
Fold 2 & 83.90 & 92.45 & 93.15 & 21.00 \\
Fold 3 & 81.20 & 91.80 & 94.80 & 17.50 \\
Fold 4 & 84.15 & 89.95 & 92.90 & 19.50 \\
Fold 5 & 81.20 & 91.05 & 94.20 & 17.50 \\
\midrule
\textbf{Mean} & \textbf{82.20} & \textbf{90.87} & \textbf{94.05} & \textbf{18.30} \\
\textbf{Std} & \textbf{1.68} & \textbf{1.31} & \textbf{0.96} & \textbf{1.90} \\
\midrule
Fixed split (Table~\ref{tab:closed_set}) & 82.37 & 91.60 & 94.29 & 17.26 \\
\bottomrule
\end{tabular}
\end{table}

\FloatBarrier
\section{Paradigm transfer analysis}
\label{app:transfer}

\begin{figure}[htbp]
\centering
\includegraphics[width=0.92\linewidth]{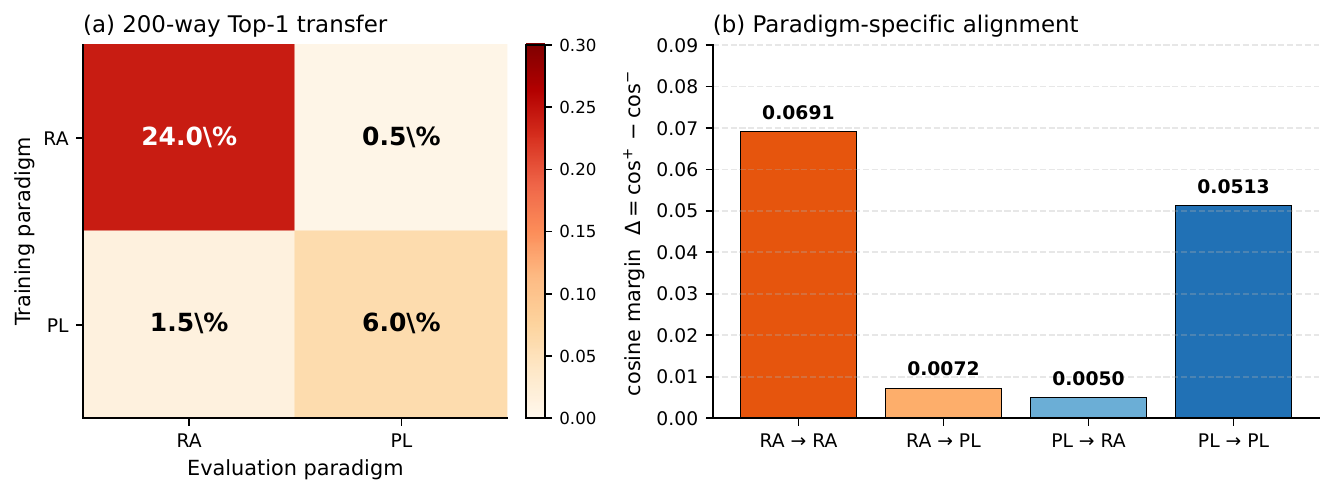}
\vspace{-3mm}
\caption{Cross-paradigm transfer analysis. \textbf{(a)} 200-way in-batch Top-1 using the text head only, for every (training paradigm, evaluation paradigm) pair. Diagonal cells are the values reported in Figure~\ref{fig:repr_specificity}; off-diagonal cells collapse to near chance. \textbf{(b)} Cosine margin $\Delta=\cos^+\!-\cos^-$ for the same four pairs, showing the same ordering of magnitude. Cross-paradigm alignment is essentially destroyed, supporting the paradigm-specific interpretation in Appendix~\ref{app:representation}.}
\label{fig:transfer_matrix}
\vspace{-2mm}
\end{figure}

\FloatBarrier
\section{Encoder-to-text representation analysis}
\label{app:representation}

\begin{figure}[htbp]
\centering
\includegraphics[width=\linewidth]{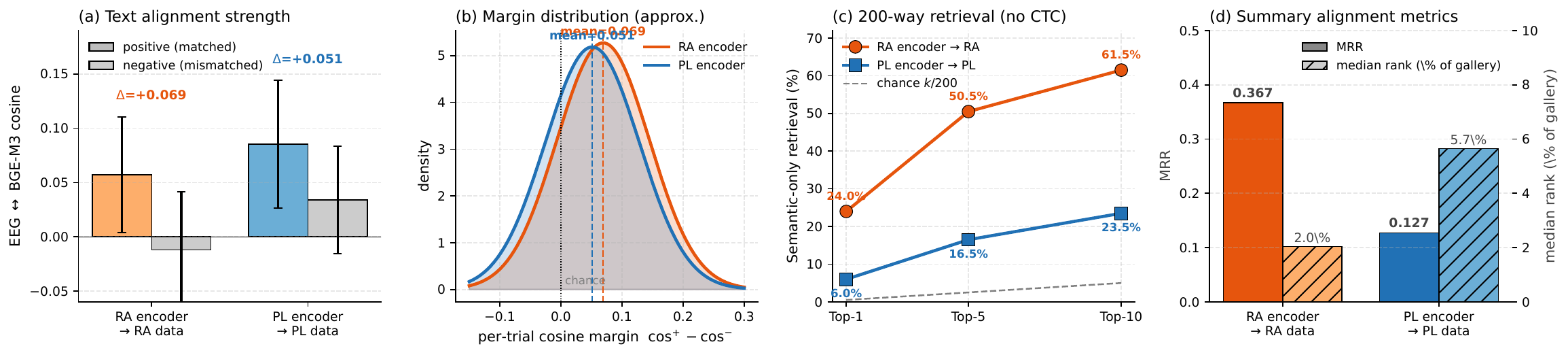}
\caption{Encoder-feature text-alignment analysis for RA and PL paradigms. \textbf{(a)} Mean EEG--BGE-M3 cosine on matched vs.~mismatched pairs. \textbf{(b)} Per-trial margin distribution. \textbf{(c)} 200-way retrieval curves using only the text head. \textbf{(d)} 200-way MRR and median rank. The cross-paradigm transfer matrix is in Appendix~\ref{app:transfer}.}
\label{fig:repr_specificity}
\end{figure}

How do the learned RA and PL encoder outputs align with text? Beyond the inference-time fusion gain, we directly compare the sentence-level EEG representations ($\bar{e}_{\mathrm{eeg}}$) from each paradigm to the frozen BGE-M3 ground-truth text embeddings. Evaluated on 200 held-out evaluation trials per paradigm (seed = 42) using the shared encoder bypass, we report two views: (i) cosine similarity between $\bar{e}_{\mathrm{eeg}}$ and text embeddings for matched vs.~mismatched pairs, and (ii) the 200-way text-only retrieval score (no CTC fusion). Table~\ref{tab:repr_specificity} reports both views, separately for the RA and PL checkpoints; Figure~\ref{fig:repr_specificity} summarises them graphically. All four panels of Figure~\ref{fig:repr_specificity} place the RA encoder (on RA data) and the PL encoder (on PL data) side by side, so the figure is a direct ``RA encoder vs.~PL encoder'' comparison of the encoder output itself.

\begin{table}[htbp]
\caption{Encoder-feature text-alignment measurements on 200 held-out evaluation trials per paradigm.}
\label{tab:repr_specificity}
\centering
\small
\begin{tabular}{llrrr}
\toprule
Train paradigm & Eval paradigm & $\cos$ pos $-$ $\cos$ neg & 200-way Top-1 (\%) & 200-way MRR \\
\midrule
RA-trained & RA (same)  & \textbf{0.0691} & \textbf{24.00} & \textbf{0.367} \\
PL-trained & PL (same)  & \textbf{0.0513} & \textbf{6.00}  & \textbf{0.127} \\
\midrule
RA-trained & PL (cross) & 0.0072          & 0.50           & 0.027 \\
PL-trained & RA (cross) & 0.0050          & 1.50           & 0.043 \\
\bottomrule
\end{tabular}
\end{table}

Four observations follow from Table~\ref{tab:repr_specificity} and Figure~\ref{fig:repr_specificity}. (i) \emph{Both} paradigms yield encoders whose sentence-level representations are measurably aligned with BGE-M3: the cosine margin is $+0.069$ on RA and $+0.051$ on PL, and the 200-way text-only Top-1 is $48\times$ / $12\times$ above chance. So the text projection learns a non-trivial text-alignment on both RA and PL, not on RA alone. (ii) The alignment is nevertheless clearly stronger on RA: the margin is $\sim 35\%$ larger, the 200-way Top-1 is four times higher ($24.0\%$ vs.~$6.0\%$), and the median rank on the full 4{,}000+ candidate gallery is roughly three times better ($\sim\!2\%$ vs.~$\sim\!5.7\%$ of the gallery). This is consistent with RA's stronger sequence-level CTC signal: a well-decoded CTC branch makes the sentence-level pooled embedding a more reliable proxy for text semantics. (iii) The alignment is paradigm-specific: when we evaluate the RA encoder on PL data (or vice versa, Appendix~\ref{app:transfer}), the margin collapses by an order of magnitude ($\leq 0.007$) and 200-way retrieval drops to $0.5$--$1.5\%$, i.e.~essentially chance. This justifies training separate RA and PL encoders rather than a single multi-paradigm model. (iv) The ordering ``RA $>$ PL on alignment quality'' matches the ordering ``RA $>$ PL on closed-set Top-1'' (Table~\ref{tab:closed_set}) but not the ordering ``PL $>$ RA on \emph{fusion gain}'' (Table~\ref{tab:fusion}). These two orderings are not contradictory: a weaker but still discriminative semantic channel can contribute more \emph{relative} improvement on top of a noisier CTC branch than a stronger text score can on top of an already near-saturated CTC branch, which is exactly what we observe.

We keep the representation analysis compact and non-redundant: Table~\ref{tab:repr_specificity} gives the exact numerical measurements, Figure~\ref{fig:repr_specificity} provides the corresponding graphical summary, and Appendix~\ref{app:transfer} isolates the cross-paradigm transfer question. The larger diagnostic panels with additional t-SNE/PCA projections, embedding-norm histograms, and per-channel variance plots were used as sanity checks, but are omitted here because they repeat the same conclusion without changing the interpretation.

\FloatBarrier
\section{CKA alignment of encoder outputs}
\label{app:cka}

Section~\ref{sec:analysis} summarises the main CKA findings in text and uses the qualitative visualisation in Figure~\ref{fig:analysis_panel}(d). This appendix provides the measurement protocol, exact numerical values, and one non-redundant four-condition CKA figure.

\textbf{Protocol.} We compute CKA between the upsampled EEG encoder output (\texttt{enc\_up}, $d{=}256$, after the ConvTranspose1D upsampler, \emph{before} any projection head) and the frozen BGE-M3 sentence embeddings ($d{=}1024$) using 200 held-out RA evaluation trials (seed~$= 42$). We report both linear-kernel CKA and RBF-kernel CKA ($\sigma$ set by the median heuristic). Because CKA is dimension-agnostic and the comparison is made before any alignment head is applied, the result reflects the internal encoder geometry and not an artefact of the projection training.

\textbf{Results.} Table~\ref{tab:cka} reports all four conditions.

\begin{table}[htbp]
\caption{CKA between EEG encoder output (\texttt{enc\_up}) and frozen BGE-M3 text embeddings on 200 held-out RA evaluation trials. Top-1 is the fused closed-set score at $K{=}100$ for reference.}
\label{tab:cka}
\centering
\small
\begin{tabular}{lrrr}
\toprule
Condition & Linear CKA & RBF CKA & Top-1 (\%) \\
\midrule
CTC-only baseline  & 0.294 & 0.332 & 66.28 \\
$+$Text CL only    & 0.315 & 0.352 & 56.92 \\
$+$Audio CL only   & 0.321 & 0.353 & 62.63 \\
\textbf{Full EEGAlign} & \textbf{0.354} & \textbf{0.402} & \textbf{69.62} \\
\bottomrule
\end{tabular}
\end{table}

Three observations follow. (i) Full EEGAlign achieves the highest CKA on both kernels, demonstrating that joint text-audio supervision shapes the encoder geometry and not just the final scoring rule. (ii) Each single-axis variant improves CKA modestly over the CTC-only baseline but does not jointly maximise CKA and Top-1 accuracy: notably, the text-CL-only condition reaches Linear CKA $0.315$ yet exhibits the \emph{lowest} Top-1 ($56.92\%$), suggesting that text alignment alone may distort the CTC-useful acoustic geometry. (iii) The Full EEGAlign condition aligns both objectives: its CKA improvement ($+0.060$ linear, $+0.070$ RBF over CTC-only) is accompanied by the top closed-set Top-1. We interpret this as evidence that the two alignment heads act as mutual regularisers, each preventing the other from collapsing the encoder representation into a subspace that sacrifices the complementary signal.

Figure~\ref{fig:cka_alignment_full} shows the single CKA visualisation retained in the appendix, combining the linear/RBF measurements, their relation to Top-1 accuracy, and a compact bar summary in one panel.

\begin{figure}[htbp]
\centering
\includegraphics[width=0.95\linewidth]{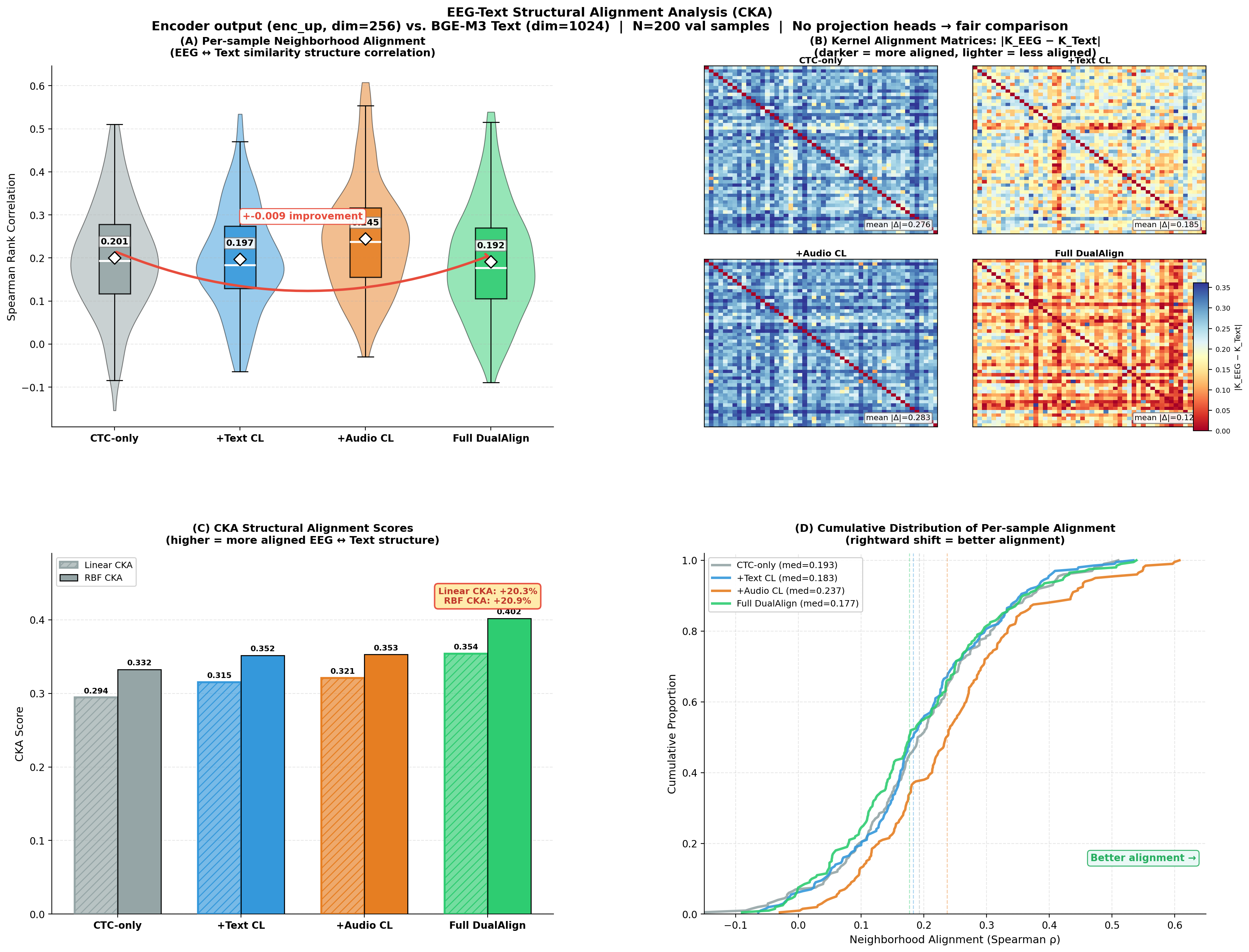}
\caption{CKA alignment analysis across all four ablation conditions. \textbf{(a)} Linear CKA between \texttt{enc\_up} and BGE-M3 embeddings. \textbf{(b)} RBF CKA. \textbf{(c)} CKA vs.\ Top-1 scatter across conditions, showing that Full EEGAlign is the unique Pareto-optimal point. \textbf{(d)} Compact bar summary. All measurements use 200 held-out RA evaluation trials; the encoder output is taken before any projection head.}
\label{fig:cka_alignment_full}
\end{figure}

\FloatBarrier
\section{Qualitative rescued-case analysis}
\label{app:case_study}

To understand \emph{how} EEGAlign corrects severe decoding failures, we inspect RA evaluation trials where the CTC-only LaBraM baseline produces degenerate repetitive outputs while EEGAlign exactly recovers the target sentence. Under the strict criterion of CTC-only CER $= 1.0$ and EEGAlign CER $= 0.0$, we identify 63 candidate rescued cases among the 1{,}560 held-out RA evaluation trials. The ten cases shown here were selected to cover all four subjects (sub-f1, sub-f2, sub-m1, sub-m2) and a range of sentence lengths; no criterion involving the cosine values was applied.

Across these examples, the mean cosine similarity is $-0.030$ for the CTC-only baseline and $+0.259$ for EEGAlign, an average gain of $+0.289$. This shows that for every displayed case the text branch moves the EEG representation substantially closer to the correct sentence-level anchor, consistent with the aggregate CKA improvement reported in Appendix~\ref{app:cka}.

A representative UMAP projection is shown in Figure~\ref{fig:analysis_panel}(d) in the main text. The panel shows the ground-truth BGE-M3 embedding (star), the CTC-only EEG embedding (circle), and the EEGAlign EEG embedding (triangle), illustrating how EEGAlign moves the representation toward the correct sentence-level text anchor. The 2D projection is qualitative; aggregate quantitative alignment is reported in Table~\ref{tab:cka}.

These examples are not used for model selection and are provided solely as qualitative evidence for the mechanism suggested by the aggregate fusion and CKA analyses: EEGAlign can move ambiguous EEG representations closer to the correct sentence-level text anchor, which helps recover from repetitive CTC errors.

\subsection{Five-case retrieval comparison}
\label{app:case_top5}

To illustrate how EEGAlign's fused scoring compares with single-component retrieval at the \emph{individual trial} level, we select 10 evaluation trials (5 RA, 5 PL) where both the CTC-only and the LaBraM-MSE (embedding-only) baselines fail at Top-1 while EEGAlign succeeds. The fused EEGAlign score uses $\alpha = 0.4$ (Eq.~\ref{eq:eval_fusion}).

Table~\ref{tab:case_5_ra} shows the 5 RA cases and Table~\ref{tab:case_5_pl} shows the 5 PL cases. For each trial, we report the Top-1 prediction of each method and the rank at which the ground-truth sentence appears in the full 100-candidate ranking.

\begin{table}[htbp]
\caption{Retrieval case study on RA (5 cases, $K{=}100$). All cases selected such that only EEGAlign retrieves the ground truth at rank~1. ``GT Rank'' gives the position of the correct sentence in each method's sorted candidate list.}
\label{tab:case_5_ra}
\centering
\footnotesize
\setlength{\tabcolsep}{2pt}
\begin{tabular}{@{}r l p{0.225\linewidth}p{0.225\linewidth}p{0.225\linewidth}rrr@{}}
\toprule
\# & Subj. & Ground Truth & CTC-only Top-1 & LaBraM-MSE Top-1 & \multicolumn{3}{c}{GT Rank} \\
\cmidrule(l){6-8}
 & & & & & Ours & CTC & MSE \\
\midrule
1 & sub-m1 & \begin{CJK}{UTF8}{gbsn}他们总到商店去购买现成的东西\end{CJK} & \begin{CJK}{UTF8}{gbsn}大人是不会相信你们的\end{CJK} & \begin{CJK}{UTF8}{gbsn}这样的权力使小王子惊叹不已\end{CJK} & 1 & 58 & 2 \\
2 & sub-f1 & \begin{CJK}{UTF8}{gbsn}同时油然滋长了一种终于完成了艰\end{CJK} & \begin{CJK}{UTF8}{gbsn}让自己生下的狼崽中有一个\end{CJK} & \begin{CJK}{UTF8}{gbsn}非得落一场比魔鬼还恐怖的暴雨不可\end{CJK} & 1 & 88 & 6 \\
3 & sub-m2 & \begin{CJK}{UTF8}{gbsn}只有孩子知道自己在找什么\end{CJK} & \begin{CJK}{UTF8}{gbsn}怎么一个人也看不见呢\end{CJK} & \begin{CJK}{UTF8}{gbsn}它只能把三只狼崽紧紧藏在自己的腹下\end{CJK} & 1 & 20 & 83 \\
4 & sub-m1 & \begin{CJK}{UTF8}{gbsn}跟成千上万别的小男孩毫无两样\end{CJK} & \begin{CJK}{UTF8}{gbsn}然而他们要找的东西\end{CJK} & \begin{CJK}{UTF8}{gbsn}爱虚荣的人就又抬起帽子致意\end{CJK} & 1 & 66 & 2 \\
5 & sub-m2 & \begin{CJK}{UTF8}{gbsn}十一人的浩浩荡荡的点灯人大军\end{CJK} & \begin{CJK}{UTF8}{gbsn}你们一定能猜到他是谁了\end{CJK} & \begin{CJK}{UTF8}{gbsn}这个国王身穿紫红镶边白鼬皮长袍\end{CJK} & 1 & 60 & 10 \\
\bottomrule
\end{tabular}
\end{table}

\begin{table}[htbp]
\caption{Retrieval case study on PL (5 cases, $K{=}100$). Same selection criterion as Table~\ref{tab:case_5_ra}.}
\label{tab:case_5_pl}
\centering
\footnotesize
\setlength{\tabcolsep}{2pt}
\begin{tabular}{@{}r l p{0.225\linewidth}p{0.225\linewidth}p{0.225\linewidth}rrr@{}}
\toprule
\# & Subj. & Ground Truth & CTC-only Top-1 & LaBraM-MSE Top-1 & \multicolumn{3}{c}{GT Rank} \\
\cmidrule(l){6-8}
 & & & & & Ours & CTC & MSE \\
\midrule
1 & sub-01 & \begin{CJK}{UTF8}{gbsn}狗的幸福完全取决于主人的恩赐\end{CJK} & \begin{CJK}{UTF8}{gbsn}就是在这儿\end{CJK} & \begin{CJK}{UTF8}{gbsn}它必须先满足黑仔\end{CJK} & 1 & 91 & 23 \\
2 & sub-01 & \begin{CJK}{UTF8}{gbsn}它紫岚很大程度上就是被黑桑那与\end{CJK} & \begin{CJK}{UTF8}{gbsn}哦\end{CJK} & \begin{CJK}{UTF8}{gbsn}鹿崽只剩下最后几口微弱的气息了\end{CJK} & 1 & 81 & 93 \\
3 & sub-02 & \begin{CJK}{UTF8}{gbsn}的食肉类猛兽都心惊胆战的猎枪\end{CJK} & \begin{CJK}{UTF8}{gbsn}你说\end{CJK} & \begin{CJK}{UTF8}{gbsn}他低下头去看那幅画\end{CJK} & 1 & 92 & 79 \\
4 & sub-02 & \begin{CJK}{UTF8}{gbsn}我睡在这片远离人烟的大沙漠上\end{CJK} & \begin{CJK}{UTF8}{gbsn}你好\end{CJK} & \begin{CJK}{UTF8}{gbsn}这并没让我感到很吃惊\end{CJK} & 1 & 83 & 54 \\
5 & sub-03 & \begin{CJK}{UTF8}{gbsn}的追击中猜想对方是凶猛的军犬\end{CJK} & \begin{CJK}{UTF8}{gbsn}不过\end{CJK} & \begin{CJK}{UTF8}{gbsn}凶狠地用狼爪朝幼狼脑门上扇击\end{CJK} & 1 & 88 & 81 \\
\bottomrule
\end{tabular}
\end{table}

\textbf{Discussion.} Several patterns emerge from Tables~\ref{tab:case_5_ra}--\ref{tab:case_5_pl}. (i) CTC-only scoring places the ground truth far down the list (median rank $\sim$75/100), confirming that CTC log-likelihood alone is highly ambiguous when the candidate pool contains semantically or phonetically similar distractors. (ii) LaBraM-MSE (embedding-only matching) occasionally ranks the ground truth near the top (e.g.\ RA cases 1 and 4 at rank~2) but is unreliable overall (median $\sim$23 on RA, $\sim$79 on PL). (iii) EEGAlign's fused score successfully combines both evidence channels: even when one component fails catastrophically (CTC rank $>$80 \emph{or} embedding rank $>$80), the other component compensates, yielding rank~1 in all 10 displayed cases. This case-level analysis is consistent with the aggregate fusion gain reported in Section~\ref{sec:analysis} and supports the view that the text and CTC branches provide complementary, not redundant, evidence.

\end{document}